\documentclass{article}

\usepackage[preprint]{neurips_2022}

\usepackage[utf8]{inputenc} 
\usepackage[T1]{fontenc}    
\usepackage{hyperref}       
\usepackage{url}            
\usepackage{booktabs}       
\usepackage{amsfonts}       
\usepackage{nicefrac}       
\usepackage{microtype}      
\usepackage{xcolor}         

\usepackage{soul}
\usepackage{graphicx}
\usepackage{subfig}
\usepackage{enumitem}
\usepackage{amsmath}
\usepackage{bm}
\usepackage{bbm}
\usepackage{multirow}

\newcommand{\ie}{\textit{i}.\textit{e}.,~}
\newcommand{\eg}{\textit{e}.\textit{g}.,~}
\newcommand{\vs}{\textit{vs}.~}

\newcommand{\fig}[1]{Figure #1}
\newcommand{\tab}[1]{Table #1}

\title{Self-Supervised Implicit Attention: Guided Attention by The Model Itself}

\author{%
  Jinyi Wu \qquad Xun Gong \qquad Zhemin Zhang \\
  Southwest Jiaotong University \\
  \texttt{watson753@my.swjtu.edu.cn} \\
}

\begin{document}

\maketitle

\sethlcolor{yellow}
\setstcolor{red}
\setcitestyle{numbers}

\begin{abstract}
  We propose \textbf{S}elf-\textbf{S}upervised \textbf{I}mplicit \textbf{A}ttention (SSIA),
  a new approach that adaptively guides deep neural network models to gain attention by 
  exploiting the properties of the models themselves.
  SSIA is a novel attention mechanism that does not require any extra parameters, computation,
  or memory access costs during inference, which is in contrast to existing attention mechanism.
  In short, by considering attention weights as higher-level semantic information,
  we reconsidered the implementation of existing attention mechanisms 
  and further propose generating supervisory signals from higher network layers to guide 
  lower network layers for parameter updates. We achieved this by building a self-supervised 
  learning task using the hierarchical features of the network itself, which 
  \textbf{only} works at the training stage.
  To verify the effectiveness of SSIA, we performed a particular implementation 
  (called an SSIA block) in convolutional neural network models and validated it 
  on several image classification datasets.
  The experimental results show that an SSIA block can significantly improve the model performance,
  even outperforms many popular attention methods that require additional 
  parameters and computation costs, such as Squeeze-and-Excitation and 
  Convolutional Block Attention Module.
  Our implementation will be available on GitHub.
\end{abstract}

\section{Introduction}

In recent years, researchers have made many efforts to solve the 
problem of training very deep neural networks, such as designing 
better optimizers \cite{Zeiler2012ADADELTAAA, Kingma2015AdamAM},
normalizing the initialization of model parameters \cite{Glorot2010UnderstandingTD, He2015DelvingDI},
using residual structures \cite{He2016DeepRL, Huang2017DenselyCC}, and so on.
As these techniques have evolved, it has become possible to train very deep networks.
These deep models, such as \cite{Szegedy2016RethinkingTI, He2016DeepRL, Chollet2017XceptionDL},
have been proven to be powerful tools in various fields \cite{Lin2014MicrosoftCC, Russakovsky2015ImageNetLS, Chen2018DeepLabSI};
however, improving model performance by increasing the model parameters and computation 
is inefficient, and making models difficult to implement in practice.
Therefore, self-attention methods that can improve model performance with 
fewer parameters and lower computation costs are becoming popular.

Existing attention methods proposed for convolutional neural network (CNN) models 
are typically designed as specialized modules, they inserted directly into the 
baseline CNN models to calculate attention weights along different dimensions of 
the feature maps \cite{Wang2017ResidualAN, Hu2018SqueezeandExcitationN, Woo2018CBAMCB,
Hu2018GatherExciteEF, Chen2018A2NetsDA, Wang2018NonlocalNN}.
These calculated attention weights are believed to guide the baseline models to focus 
on \emph{what} and \emph{where}, which is \textbf{manually} applied to the input features 
to produce a refined feature map with element-wise
multiplication, as in Squeeze-and-Excite (SE) \cite{Hu2018SqueezeandExcitationN} and 
Convolutional Block Attention Module (CBAM) \cite{Woo2018CBAMCB},
or addition, such as in the Non-local Neural Networks (NLN) \cite{Wang2018NonlocalNN} and 
Global Context Network (GCNet) \cite{Cao2019GCNetNN}.
These attention modules typically do not require additional supervision and are 
trained together with the baseline models to learn the distribution of the weights 
likely to identity useful features.
We argue that these attention modules act on the feature maps 
explicitly and that their training is driven only by the weak supervisory signals 
generated from the top layer of the network with task-related annotations (\eg category labels).
We call this type of attention weakly supervised explicit attention.
Such attention methods generally require additional parameters and computational costs. 
Although several attention methods requiring only a few (or even no) parameters 
or computational costs have been proposed recently \cite{Wang2020ECANetEC, Yang2021SimAMAS},
they, as explicit attention, still require the inferring of attention weights and 
refining of feature maps.
This increases the memory access and computational process, and hence reducing 
the inference speed.

In this paper, we proposed 
\textbf{S}elf-\textbf{S}upervised \textbf{I}mplicit \textbf{A}ttention (SSIA)
and its particular implementation, the SSIA block. SSIA is a fundamentally different 
attention mechanism that requires no additional parameters, computation, or memory access 
costs during inference. The core idea of implementing SSIA is to use 
the model's intermediate feature maps to guide the attention of the model itself.
By exploiting the inherent nature of deep neural networks during the 
training phase, supervisory signals, which guide parameter updates in 
the low network layers, are generated from the intermediate outputs of 
the networks. Since the SSIA implementation does \textbf{not} require 
refining feature maps explicitly but rather guides the baseline model 
to learn feature representations with better attentional properties 
during training, it does not need to calculate attention weights explicitly 
at the inference phase.
In this aspect, SSIA seems to behave more like a regularization approach,
such as weight decay \cite{Krogh1991ASW} or dropout \cite{Srivastava2014DropoutAS}.
Probably the studies most closely related to ours are the recent works 
\cite{Linsley2019LearningWA, Fukui2019AttentionBN, Li2020DeepRA, Patro2021SelfSF} 
attempting to train attention modules through more direct supervised signals,
such as extra annotation information \cite{Linsley2019LearningWA} 
or reinforcement learning \cite{Li2020DeepRA}.
However, they still follow the idea of the mainstream explicit attention methods,
which consider the attention module to be a specialized module functionally separated from 
the baseline model and implement attention by manually refining feature maps.
In contrast, our SSIA builds self-supervised learning tasks from the baseline model itself 
by leveraging the \emph{feature hierarchical nature} 
\cite{Zeiler2014VisualizingAU, Yosinski2015UnderstandingNN, Goodfellow2015DeepL},
and gives the low layer features better attention properties by learning.
Because SSIA does not work by explicitly modifying feature maps,
we argue that it imposes \emph{implicit} attention on baseline models.



In summary, our main contributions are as follows:
\begin{enumerate}
   \item We reconsidered the implementation of existing attention mechanisms in 
   deep learning models and propose a self-supervised attention training technique.
   The result is SSIA, a novel attention mechanism.
   \item We designed a novel attention module, the SSIA block, to implement 
   the proposed SSIA mechanism on computer vision tasks and study its advantages.
   \item We validate the SSIA block on several classification datasets,
   which showed that it significantly improves the model performance without 
   requiring extra parameters, computation, or memory access costs during inference.
\end{enumerate}

\section{Method}

\subsection{Motivation}
\label{motivation}

\begin{figure}[tb]
  \centering
  \subfloat[Normal attention]{
     \includegraphics[width=0.48\linewidth]{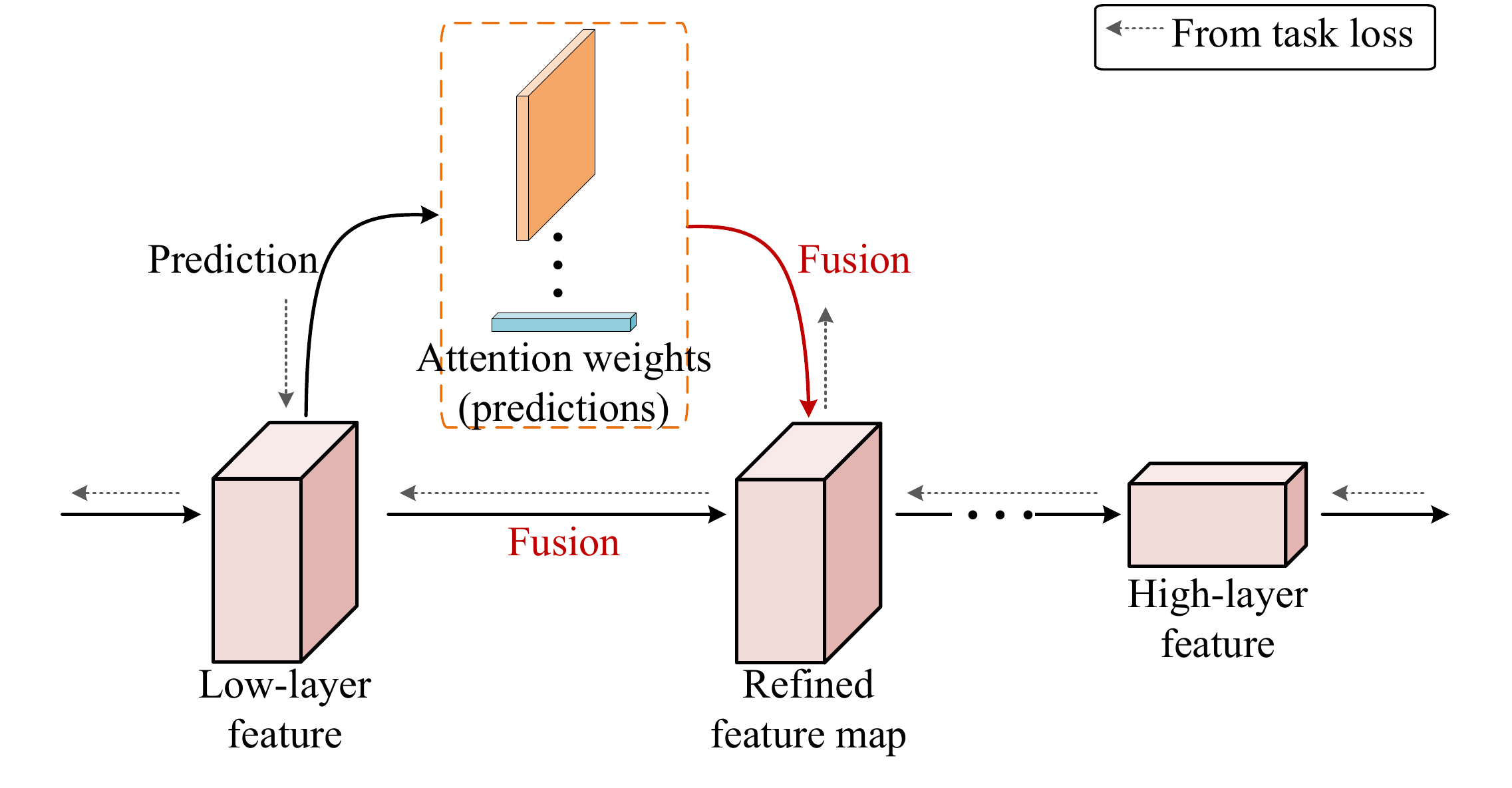}
     \label{fig:figure1:a}
  }
  \subfloat[SSIA attention]{
     \includegraphics[width=0.48\linewidth]{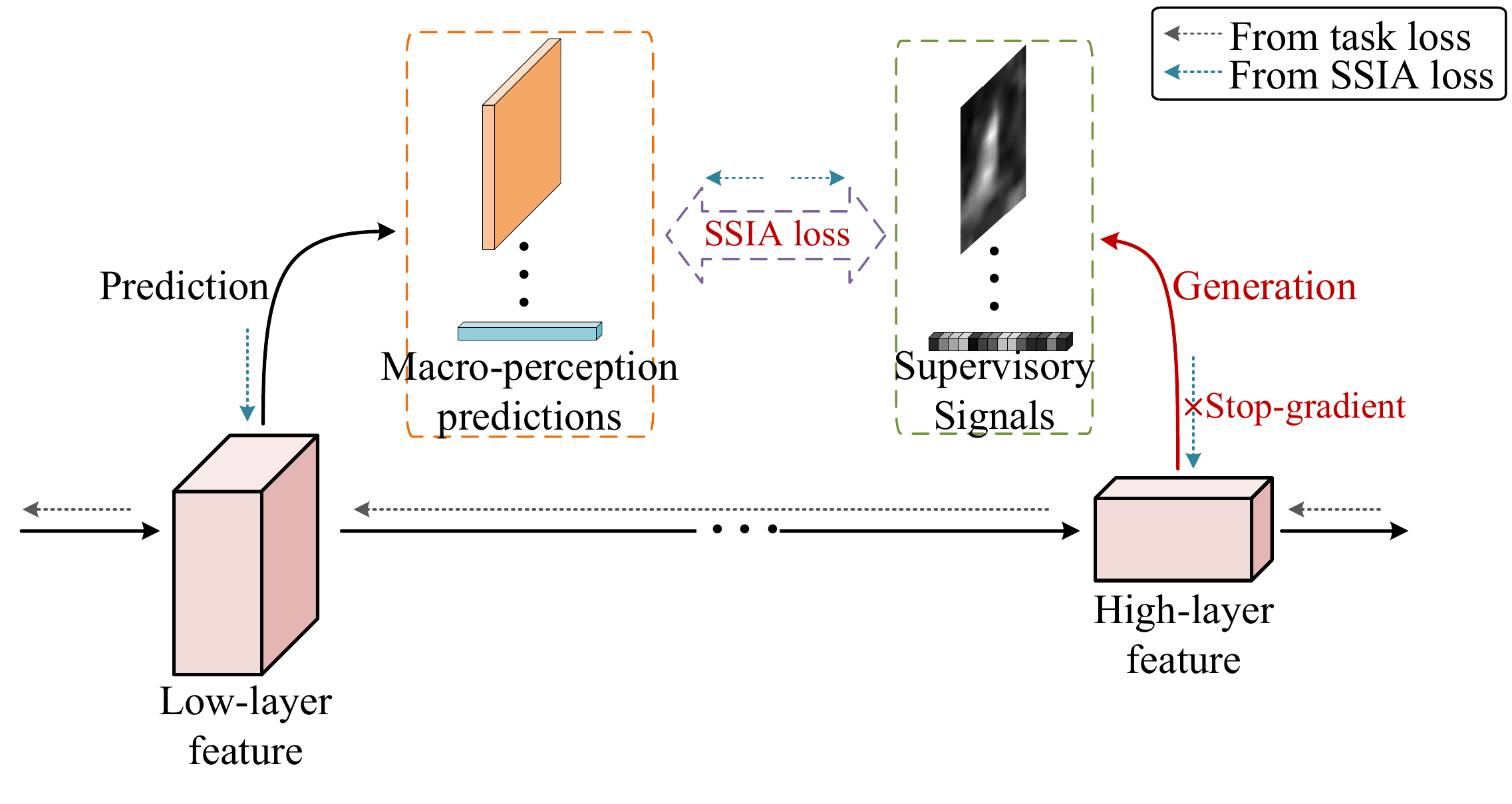}
     \label{fig:figure1:b}
  }\\
  \caption{Comparison on SSIA attention and other attention methods. 
  (a) is generalized "explicit" attention, which implements 
  attention by refining feature maps, and (b) is our "implicit" SSIA,
  which is guided by the built self-supervised task losses.
  The red color is used in the figures to highlight their differences.
  The gray dashed arrows indicate the top-down gradient flow of the target task loss,
  and the blue dashed arrows indicate the gradient flow of the SSIA loss.}
  \label{fig:figure1}
\end{figure}

Fundamentally, we know that the essence of attention mechanism is 
the perception of global-level information and its use in 
the assignment of computational weights. Thus, the models with 
attention can retain more task-relevant valuable information at a (relatively) 
more global level in the low-to-high forward-propagation process. We considered 
this definition of attention and made the following two observations:

\begin{itemize}[left=0em]
   \item \emph{First}, we observed that previous attention approaches 
   \cite{Hu2018SqueezeandExcitationN, Woo2018CBAMCB, Cao2019GCNetNN,
   Park2020ASA, Wang2020ECANetEC, Yang2021SimAMAS} could be generally 
   divided into three steps: (1) modeling the context of the 
   input features to generate global descriptors, (2) performing nonlinear 
   transformations on global descriptors to generate attention weights, (3) 
   refining the input feature maps by applying these attention weights.
   The pipeline is shown in \fig{\ref{fig:figure1:a}}. We argue that these 
   attention methods actually perform the perception of global-level information 
   in Steps 1 and 2, and perform the assignment of computational weights 
   in Step 3.
   \item \emph{Second}, according to previous studies 
   \cite{Zeiler2014VisualizingAU, Yosinski2015UnderstandingNN}, features learned by 
   deep convolutional neural networks (DCNNs) have a self-organized and hierarchical 
   feature representation, that is, a \emph{hierarchical-feature nature} 
   \cite{Goodfellow2015DeepL}. This means that DCNN models produce feature maps with 
   higher-level semantic representations at the higher network layers (in both 
   spatial and channel dimensions), which naturally have a more global-level perception 
   than lower network layers.
\end{itemize}

Combining these two observations, we reexamined previous implementations of 
the attention mechanism and concluded that the three-steps process, as shown 
in \fig{\ref{fig:figure1:a}}, is \textbf{not} necessary --- it is not necessary 
to generate attention weights explicitly and assign computational weights \emph{manually}.
Attention can also be implemented in a neural network model if the original 
low-layer features can acquire the perception of global-level information 
(denoted as macro-perception) through the training process,
allowing subsequent network layers to perform \emph{adaptive} 
assignment of computational weights.
Combined with the hierarchical-feature nature of CNNs, we posed an intuitive question: 
can low network layers benefit from the natural macro-perception of higher 
network layers during training, thus guiding the model to gain attention? 

Based on the above idea, we devised our novel attention mechanism, SSIA,
and built a particular implementation called the SSIA block for typical CNN models.
As shown in \fig{\ref{fig:figure1:b}}, the SSIA block attempts to build a self-supervised 
learning task that enables the features learned in lower network layers to predict 
higher-level features using a weak predictor and penalizing prediction failures.
To reduce the prediction error, the trained low network layers tend to produce features 
with more high-level semantic properties, which the subsequent network layers can use.
The SSIA block can be discarded after training, whereas the SSIA attention is implemented in the model.


\subsection{SSIA Block}

SSIA is implemented on CNNs by a computational unit called an SSIA block,
which is plugged into the baseline model during the training phase.
Each SSIA block accepts two different layers of feature maps as input 
and yields an auxiliary self-supervised loss.
Specifically, each SSIA block contains three main parts:
(i) a \emph{weak predictor}, the macro-perception predictor (MPP), which predicts 
the macro-perception information that the input low-layer features should 
have when they forward-propagate to the higher layers,
(ii) an untrainable \emph{supervisory signal generator}, which uses 
high-layer features to generate supervisory signals, and 
(iii) an \emph{SSIA loss} which measures the deviation of the predictions. 

\begin{figure}[tb]
   \centering
   \includegraphics[width=0.8\linewidth]{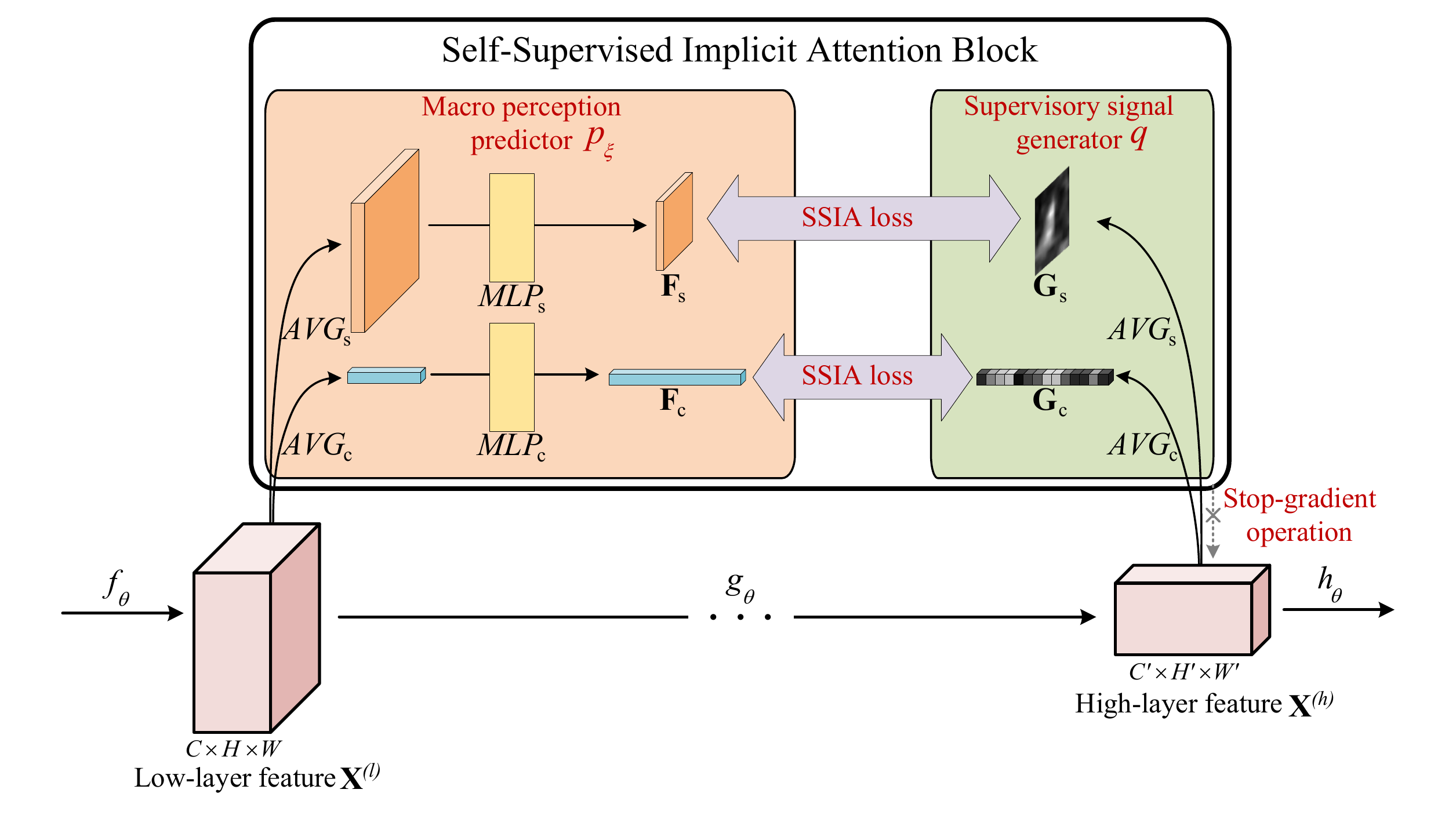}\\
   \caption{SSIA block architecture. It's a particular implementation of 
   SSIA on CNNs, it takes feature maps of two different depth layers as inputs to 
   generate macro-perception predictions and supervisory signals, respectively,
   then computes SSIA losses in spatial and channel dimensions.}
   \label{fig:figure2}
\end{figure}

The workflow of an SSIA block is shown in \fig{\ref{fig:figure2}}. 
These three components work together to guide the MPP and low network 
layers in updating.
For simplicity, we denote the input side of the MPP in the SSIA block as the 
\emph{prediction side} and the input side of the supervisory signal generator 
as the \emph{signal side}.
The backward gradients are only passed at the prediction side and 
stop gradients at the signal side.

Since MPP is a \textbf{weak} predictor, it is inherently limited in its capability.
Therefore, to reduce the self-supervised loss introduced by the SSIA block 
during back-propagation, the network tends to learn feature representations that 
have significant macro-perception semantics so that the weak MPP can easily use this 
information to make good predictions. Finally, the network outputs intermediate 
feature maps with better macro-perception information (\ie better attentional properties),
which the higher layers can further use.

More formally, let the input of the baseline model be $I$, forward-propagated on 
the network to obtain the $l$-th layer feature map 
$\mathbf{X}^{(l)} \in \mathbb{R}^{C \times H \times W}$, which can be summarized as
$\mathbf{X}^{(l)} = f_{\theta}(I)$.
Then, $\mathbf{X}^{(l)}$ continues to forward-propagate to the $h$-th layer feature map 
$\mathbf{X}^{(h)} \in \mathbb{R}^{C^{\prime} \times H^{\prime} \times W^{\prime}}$,
this process can be summarized as $\mathbf{X}^{(h)} = g_{\theta}(\mathbf{X}^{(l)})$.
We feed $\mathbf{X}^{(l)}$ and $\mathbf{X}^{(h)}$ into the prediction side and 
the signal side, respectively, of an SSIA block. Due to the hierarchical-feature nature 
of the neural network, the high-layer feature map $\mathbf{X}^{(h)}$ finds information 
on a more global level than the lower-layer feature map $\mathbf{X}^{(l)}$.
Using these notational conventions, we describe the details of each part of 
our SSIA block implementation in the following.

\paragraph{Supervisory Signal Generator.}
To guide low network layers with high-level features, the supervisory signal generator 
takes the high-layer feature map $\mathbf{X}^{(h)}$ as input and generates different types 
of the supervisory signals $\mathbf{G_1},\mathbf{G_2},\dots,\mathbf{G_m}$.
These supervisory signals can be considered as the descriptions of different aspects of 
the macro-perception information, which the high-layer feature map $\mathbf{X}^{(h)}$ 
naturally has. This process can be expressed as
$ \mathbf{G_1}, \mathbf{G_2}, \dots, \mathbf{G_m} = q(\mathbf{X}^{(h)}) $,
where $q( \cdot )$ denotes the supervisory signal generator, and $\mathbf{G_1},
\mathbf{G_2}, \dots, \mathbf{G_m}$ depend on the types of implicit attention 
to be imposed.
We only generate spatial supervisory signal 
$\mathbf{G_s} \in \mathbb{R}^{1 \times H^{\prime} \times W^{\prime}}$ and channel 
supervisory signal $\mathbf{G_c} \in \mathbb{R}^{C^{\prime} \times 1 \times 1}$
of the high-layer feature map $\mathbf{X}^{(h)}$.

In our SSIA block implementation, $q( \cdot )$ first generates the spatial and channel 
global descriptors by using global pooling operations, then normalize them 
in the corresponding dimensions to obtain $\mathbf{G_s}$and $\mathbf{G_c}$.
This is more formally expressed as:%
\begin{align}
  \mathbf{G_s} &= \mathit{Norm}( \mathit{AVG_s}( \mathbf{X}^{(h)} ) ) \text{,} \\
  \mathbf{G_c} &= \mathit{Norm}( \mathit{AVG_c}( \mathbf{X}^{(h)} ) ) \text{,}
\end{align}
where $\mathit{Norm}(\cdot)$ denotes the normalization operation applied to tensors 
(\ie subtract their mean and divide the variance). $\mathit{AVG_s}$ and $\mathit{AVG_c}$ 
denote global average pooling operations along the spatial and channel 
dimensions, respectively.

\paragraph{Macro-Perception Predictor.}
The MPP takes the low-level feature $\mathbf{X}^{(l)}$ of an image as input 
and makes predictions $\mathbf{F_1}, \mathbf{F_2}, \dots, \mathbf{F_m}$ 
about the macro-perception information it propagates to the high-level feature map 
$\mathbf{X}^{(h)}$. This process can be expressed as 
$ \mathbf{F_1}, \mathbf{F_2}, \dots, \mathbf{F_m} = p_{\xi}(\mathbf{X}^{(l)}) $,
where $p_{\xi}$ denotes the weak predictor MPP with the trainable 
parameters $\xi$. The predicted macro-perception information,
$\mathbf{F_1}, \mathbf{F_2}, \dots, \mathbf{F_m}$, corresponds to the 
supervisory signals $\mathbf{G_1},\mathbf{G_2},\dots,\mathbf{G_m}$.
Therefore, we only predict the macro-perception information 
$\mathbf{F_s} \in \mathbb{R}^{1 \times H^{\prime} \times W^{\prime}}$ and 
$\mathbf{F_c} \in \mathbb{R}^{C^{\prime} \times 1 \times 1}$ related to 
the spatial and channel dimensions, respectively, with the same tensor sizes 
as the supervisory signals $\mathbf{G_s}$ and $\mathbf{G_c}$.
The predictions $\mathbf{F_s}$ and $\mathbf{F_c}$ are analogous to the 
spatial attention (maps) and channel attention (vectors) in existing attention 
methods (\eg CBAM \cite{Woo2018CBAMCB} and BAM \cite{Park2020ASA}), which we call 
spatial macro-perception prediction and channel macro-perception prediction.

In our SSIA block implementation, $\mathbf{F_s}$ and $\mathbf{F_c}$ are computed 
as follows: \textit{1)} obtain the spatial descriptor $\bm{\phi_s}$ and channel descriptor 
$\bm{\phi_c}$ using global average pooling operations along different dimensions of the 
low-layer feature map $\mathbf{X}^{(l)}$ to aggregate statistical information,
then \textit{2)} (\textbf{optional}) normalize $\bm{\phi_s}$ and $\bm{\phi_c}$ 
and scale the spatial descriptor to reduce the computational cost, and finally,
\textit{3)} forward $\bm{\phi_s}$ and $\bm{\phi_c}$ to their respective 
multilayer perceptron (MLP) predictors to produce the macro-perception 
predictions $\mathbf{F_s}$ and $\mathbf{F_c}$, respectively.
The overall process can be expressed as follows:%
\begin{align}
   \mathbf{F_s} &= \mathit{MLP_s}(\bm{\phi_s}) = \mathit{MLP_s}(\mathit{AVG_s}(\mathbf{X}^{(l)})) \text{,} \\
   \mathbf{F_c} &= \mathit{MLP_c}(\bm{\phi_c}) = \mathit{MLP_c}(\mathit{AVG_c}(\mathbf{X}^{(l)})) \text{,}
\end{align}
where both $\mathit{MLP_s}$ and $\mathit{MLP_c}$ are MLPs 
with a single hidden layer (with a batch normalization layer). We compress 
the hidden layer size of $\mathit{MLP_s}$ and $\mathit{MLP_c}$ to a lower dimension 
to reduce their parameters and computational costs during training.
Note that the formula description above ignores, for simplicity, the optional normalization 
of $\bm{\phi_s}$ and $\bm{\phi_c}$ and the scaling of $\bm{\phi_s}$.
In practice, the spatial descriptor $\bm{\phi_s}$ can be scaled by bilinear 
interpolation or local avgrage pooling, which reduces the parameters and 
computational costs of $\mathit{MLP_s}$.

\paragraph{SSIA Loss.}
The total training loss is the weighted summation of the total SSIA loss,
$\mathcal{L}^{sb}_{total}$, and the baseline model's task loss,
$\mathcal{L}^{task}_{total}$, more formally describe as
$\mathcal{L} = \lambda^{task} \mathcal{L}^{task}_{total} + \lambda^{sb} \mathcal{L}^{sb}_{total}$,
where $\lambda^{task}$ and $\lambda^{sb}$ are hyperparameters of the relative 
contribution of these two losses. 
The total SSIA loss generated by all $N$ inserted SSIA blocks is 
$\mathcal{L}^{sb}_{total}$, which finally guides the model's attention 
during the back-propagation process. It is calculated as follows:%
\begin{equation}
   \mathcal{L}^{sb}_{total} = {\sum_{n = 1}^{N} {\lambda_{n}^{sb} \mathcal{L}^{sb} 
   \left( {\mathbf{X}^{(l_n)}, \mathbf{X}^{(h_n)}} \right)}} \text{,}
\end{equation}
where $\mathcal{L}^{sb} \left( {\mathbf{X}^{(l_n)}, \mathbf{X}^{(h_n)}} \right)$
denotes the self-supervised loss introduced by the $n$-th SSIA block,
and $\mathbf{X}^{(l_n)}$ and $\mathbf{X}^{(h_n)}$ 
denote the input feature maps of the $n$-th SSIA block's prediction and signal 
sides, respectively, which are the intermediate feature maps output from the 
baseline model. $\lambda_{n}^{sb}$ is a hyperparameter indicating the 
relative contribution of the $n$-th SSIA block.

The loss of a single SSIA block is given by
$
\mathcal{L}^{sb} \left( {\mathbf{X}^{(l_n)}, \mathbf{X}^{(h_n)}} \right) = 
{\sum_{i = 1}^{m} {\lambda^{ssia}_{i} \mathcal{L}^{ssia} 
\left( {\mathbf{F}_i, sg\left( \mathbf{G}_i \right)} \right)}}
$,
where $sg(\cdot)$ denotes the stop-gradient operation, $\mathcal{L}^{ssia}$ 
is the SSIA loss measure for each predicted macro-perception,
and $\lambda^{ssia}_{i}$ is the hyperparameter.
We only use the supervisory signals in spatial and channel dimensions,
so the self-supervised loss can be simplified to:%
\begin{equation}
\mathcal{L}^{sb} \left( {\mathbf{X}^{(l_n)},\mathbf{X}^{(h_n)}} \right) 
= \lambda_s \mathcal{L}^{ssia} ( {\mathbf{F_s},\mathbf{G_s}} )
+ \lambda_c \mathcal{L}^{ssia} ( {\mathbf{F_c},\mathbf{G_c}} ) \text{.}
\end{equation}

The SSIA loss function $\mathcal{L}^{ssia}$ measures the deviation between 
prediction results $\mathbf{F_1}, \mathbf{F_2}, \dots, \mathbf{F_m} = p_{\xi}(\mathbf{X}^{(l)})$ 
and their corresponding supervisory signals $\mathbf{G_1}, \mathbf{G_2}, \dots,
\mathbf{G_m} = q(\mathbf{X}^{(h)})$. It is related to how the MPP prediction task 
is constructed.
We constructed the MPP prediction task through regression to leverage
the supervisory signals and guide the parameter updates. Thus, $\mathcal{L}^{ssia}$
is calculated as follows:%
\begin{equation}
  \mathcal{L}^{ssia} \left( {\mathbf{F},\mathbf{G}} \right) = 
  \sum_{k} { \frac{valid \left( {\mathbf{G} (k)} \right)}
  {\varepsilon + {\sum_{k} {valid \left( {\mathbf{G} \left( k \right)} \right) }}}
  \cdot \left( {\mathbf{F} (k) - \mathbf{G} (k)} \right) ^ 2 } \text{,}
\end{equation}
where $\mathbf{F}$ indicates the MPP macro-perception prediction result of 
(for us, $\mathbf{F_s}$ and $\mathbf{F_c}$), and $\mathbf{G}$ indicates the 
supervisory signal corresponding to it (for us, $\mathbf{G_s}$ and $\mathbf{G_c}$).
The variable $k$ traverses all spatial locations when $\mathbf{G}=\mathbf{G_s}$;
similarly, the variable $k$ traverses all channels when $\mathbf{G}=\mathbf{G_c}$.
$\varepsilon$ is a small positive real number that serves to prevent 
division errors (\eg $\varepsilon = 10^{-8}$).
The function $valid \left( \cdot \right)$ is used to limit the 
supervisory signals to a specific intensity range. Specifically,%
\begin{equation}
  valid \left( \mathbf{G} (k) \right) = 
  \mathbb{I} \left[ {\left| \mathbf{G} (k) \right|} > \eta \right] \cdot
  \mathbb{I} \left[ {\left| \mathbf{G} (k) \right| < 10} \right] \text{,}
\end{equation}
where $\mathbb{I} [ \cdot ]$ is an indicator function that takes the value 
$1$ when the condition inside the square brackets is true and $0$ otherwise.
On the right side of the above equation, the first term ignores the effect 
of ambiguous supervisory signals (those close to 0), and the second term
limits the intensity of valid supervisory signals to the range 
$\left( {-10,10} \right)$, avoiding extreme loss values caused by outliers.
$\eta$ is the supervisory signal threshold for both positive and negative 
feature activation.

For convenience, we refer to the implicit attention imposed by the SSIA loss 
in the spatial (using $\mathbf{F_s}$ and $\mathbf{G_s}$) 
and channel (using $\mathbf{F_c}$ and $\mathbf{G_c}$) 
dimensions as spatial SSIA and channel SSIA, respectively.
Note that, although we used the same SSIA loss function, $\mathcal{L}^{ssia}$, for 
\textbf{all} types of macro-perception predictions, $\mathbf{F}_i$,
for similarity, it is possible to design different SSIA losses for 
each.

\section{Experiments}
\label{experiments}

In this section, we first perform a series of ablation experiments on the 
classification task over the mini-ImageNet dataset \cite{Vinyals2016MatchingNF} 
to prove the effectiveness of our proposed SSIA and analyze the design 
choices of the SSIA block. Next, we compare the performance to that of other state-of-the-art 
attention methods on the CIFAR datasets \cite{Krizhevsky2009LearningML} and ImageNet-1K dataset \cite{Russakovsky2015ImageNetLS}.
Finally, we perform a visual analysis of the baseline model trained with the SSIA block.
To evaluate the performance of the SSIA block fairly, all our experiments 
were implemented in the PyTorch framework \cite{Paszke2019PyTorchAI} 
with the same backbone, ResNet \cite{He2016DeepRL}, as the baseline model.
For other attention modules, we use the hyperparameter settings in 
their papers.

\subsection{Implement Details}

It is worth noting that the primary purpose of our experiments was to validate 
the effectiveness of the SSIA as an attention mechanism, not to find the specific 
best-performing implementation of an SSIA block.
Moreover, the best hyperparameter settings and connection positions for the SSIA block 
may be different for specific tasks and baseline models.
However, based on experimental experience, these minor differences 
in hyperparameter settings usually do \textbf{not} significantly influence the performance 
of the SSIA block. In fact, an appropriately placed SSIA block nearly always works on 
any baseline model.
Therefore, unless otherwise noted, the hyperparameters and prediction side 
placement described below were used for the SSIA block in all the experiments.

\begin{figure}[tb]
  \centering
  \includegraphics[width=0.85\linewidth]{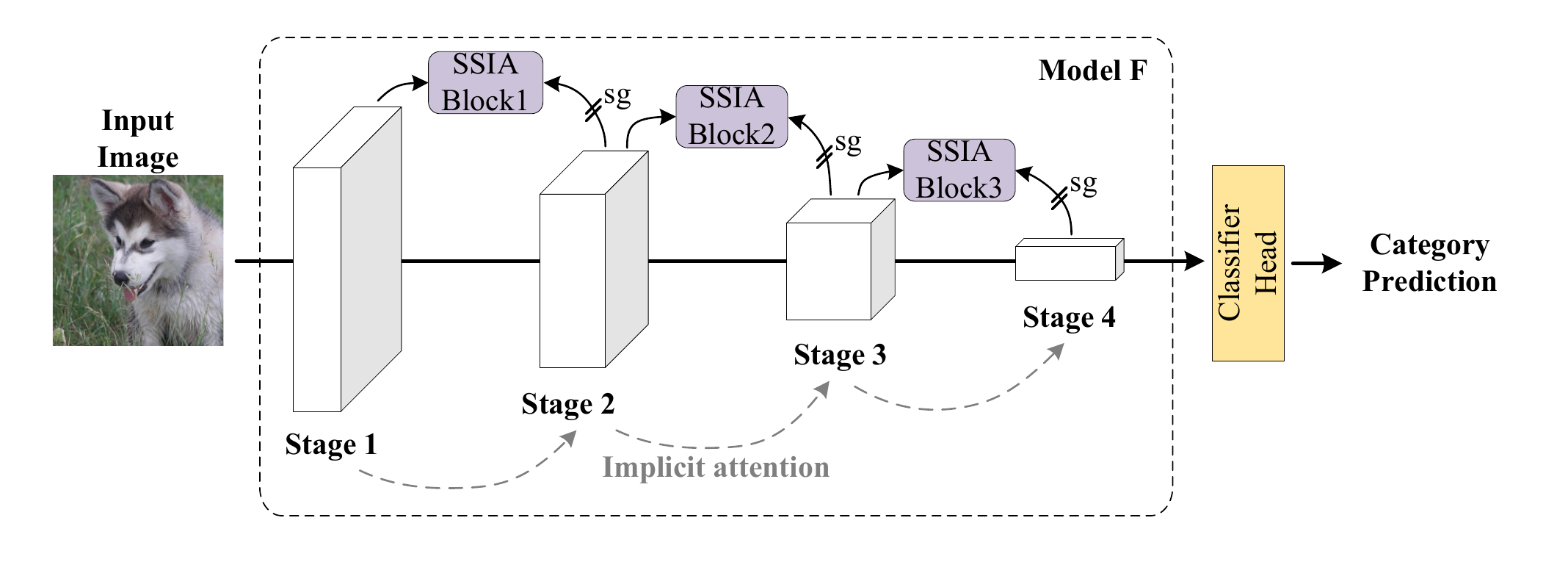}\\
  \caption{SSIA block integrated with a general ResNet model. 
  As illustrated, we plugged the SSIA block into each two adjacent stages,
  where the feature map of the lower stage is used as the prediction 
  side and the feature map of the higher stage is used as the signal side.
  "sg" means stop-gradient operation.
  The direction of the guided attention is indicated by the dashed gray arrow.}
  \label{fig:figure4}
\end{figure}

Regarding connecting the SSIA block in the baseline model, both the prediction side 
and signal side of the SSIA block can receive any size of feature maps as input,
so, \textbf{theoretically}, it can be plugged into any position in a model.
For example, with the typical ResNet \cite{He2016DeepRL} as the baseline model,
we created only three SSIA blocks, and we plugged the output feature maps of 
stages 1-3 into their prediction side,
while the output feature maps of each stage were plugged into their signal sides.
As the example shown in \fig{\ref{fig:figure4}}, each SSIA block takes the 
output feature maps of the next stage of its prediction side location as the 
signal side inputs.

Regarding the hyperparameters settings, we used the same settings for all inserted SSIA blocks.
Specifically, to reduce the parameters and computational costs, we set the 
bottleneck layer size to $d=64$ and applied 
bilinear interpolation to reduce the size of the spatial descriptors (if needed).
Note that different downsampling approaches may lead to slightly different results.
The hyperparameters that control the relative contributions of spatial and 
and channel SSIA losses are $\lambda_s=1$ and $\lambda_c=3$; the positive/negative 
example threshold is $\eta = 0.5$.
The relative contribution coefficients are $\lambda^{sb}_1 = 1$,
$\lambda^{sb}_2 = 2$, and $\lambda^{sb}_3 = 3$ (numbering from the lowest layer to highest layer),
and the contribution ratios between the total SSIA loss and the classification task loss are 
$\lambda^{task} = 1 $ and $\lambda^{sb} = 0.2$.

All models in our experiments followed the standard training pipeline.
Specifically, the 224$\times$224 image used for training was a randomly scaled cropped 
version of the original image or its horizontal mirror image. During the evaluation,
it was a centrally cropped 224$\times$224 image after scaling the shortest edge of the 
image to 256 pixels (preserving the aspect ratio).
We used an stochastic gradient descent (SGD) optimizer with a momentum of 0.9 
and a weight decay factor of 4e-5, the learning rate starting from 0.1 with 
cosine descent as in \cite{He2019BagOT}.
Without extra declaration, all networks were trained on a single GPU with 
a mini-batch size of 32 and 100 epochs totally.

\subsection{Ablation Studies}

In this subsection, we used ResNet-50 \cite{He2016DeepRL} as the baseline model 
and performed ablation experiments over the mini-ImageNet \cite{Vinyals2016MatchingNF} dataset.
The hyperparameters settings and connection positions of the SSIA blocks were 
the same as described above unless otherwise specified.
Mini-ImageNet is a subset of the ImageNet dataset, which contains 60,000 
images in 100 categories, with 600 images per category.
The original mini-ImageNet dataset is used for few-shot learning \cite{Vinyals2016MatchingNF}
with different training and testing categories.
We reintegrated it into a classification task dataset, took 20\% of the images from 
each category as the validation set, and used the remaining images as the training set.
The classification accuracy reported is for the validation set.

\begin{table}[b]
  \begin{minipage}[t]{0.47\columnwidth}
    \caption{Top-1 and Top-5 accuracies (\%) for the ResNet-50 baseline model 
    with different variants of the SSIA block on mini-ImageNet.}
    \label{tab:table1}
    \centering
    \begin{tabular}{l|c|c}
       \toprule
       Model                      & Top-1     & Top-5 \\
       \midrule
       ResNet50 (baseline)         & 79.59          & 93.45 \\
       ~+~SSIA Ch.                 & 81.54          & 94.62 \\
       ~+~SSIA Sp.                 & 81.97          & \textbf{94.67} \\
       ~\textbf{+~SSIA Ch. \& Sp.} & \textbf{82.31} & 94.65 \\
       \bottomrule
    \end{tabular}
  \end{minipage}
  \hspace{6pt}
  \begin{minipage}[t]{0.47\columnwidth}
    \caption{Top-1 and Top-5 accuracies (\%) for different signal side 
    choice schemes of the SSIA block on mini-ImageNet.}
    \label{tab:table2}
    \centering
    \begin{tabular}{l|c|c}
       \toprule
       Model                    & Top-1     & Top-5 \\
       \midrule
      ResNet50 (baseline)        & 79.59          & 93.45 \\
       ~+~final SSIA             & 82.19          & 94.61 \\
       ~\textbf{+~cascaded SSIA} & \textbf{82.31} & \textbf{94.65} \\
       ~+~identity SSIA          & 80.23          & 94.11 \\
       \bottomrule
    \end{tabular}
  \end{minipage}
\end{table}

\paragraph{Spatial SSIA \vs Channel SSIA.} We experimentally verified that both 
channel SSIA and spatial SSIA could effectively improve model performance.
\tab{\ref{tab:table1}} shows their comparative results, where 
"Ch." or/and "Sp." indicates the type of SSIA imposed (Channel or/and Spatial).
These experiments used the same hyperparameters settings and training pipeline 
except that, when only spatial SSIA was used, we set 
$\lambda_s = 1$ and $\lambda_c = 0$; when only channel SSIA was used, we set 
$\lambda_s = 0$ and $\lambda_c = 3$; when both were used, we set 
$\lambda_s = 0.5$ and $\lambda_c = 1.5$.

The experimental results show that SSIA in both the spatial and channel dimensions can 
significantly improve the accuracy of the baseline model, and their combination further 
improves the accuracy.
This suggests that the attention guided by different supervisory signals 
has different properties: they appear to guide different aspects of the 
global-level perceptibility of the baseline models.
Therefore, we use spatial SSIA and channel SSIA simultaneously in all subsequent experiments.

\paragraph{Signal Side Selection.} As previously described, the SSIA block 
can use the feature maps of any layer as the signal side input and produce 
supervisory signals, we experimentally compare the impact of this difference 
in choice on performance.
Intuitively, the higher the feature map is, the more semantic (and more global-level) 
information it contains, but also the more difficult it is to predict from 
low-layer features.
Because there are too many options for choosing the signal side location,
we only show several representative schemes, each SSIA block is connected to 
the signal side with: (i) the output feature maps of the last stage,
(ii) the output feature maps of the next stage of its prediction side location 
(\fig{\ref{fig:figure4}}), and 
(iii) the output feature maps of the same location as the prediction side.
In this experiment, the selection of the prediction sides and other settings 
were kept the same. In addition, we scaled the spatial supervisory signals of 
connected SSIA blocks to 28$\times$28, 14$\times$14, and 7$\times$7 
(from low to high) in all schemes for a fair comparison.
The experimental results are shown in \tab{\ref{tab:table2}},
where the final SSIA, cascade SSIA, and identity SSIA correspond to 
schemes (i), (ii), and (iii) described above, respectively.

The experimental results show that scheme (ii) is 
the best choice, followed by scheme (i). While scheme (iii), which uses the 
same layer feature maps as the signal side, performed significantly poorer 
than the other schemes because it does not introduce any more global-level 
perceptual information than the prediction side.
In fact, our further studies proved that the unexpected improvement 
of scheme (iii) came from some unexpected regularization effects of 
SSIA loss in the initial training stage. To eliminate this phenomenon
just ignore the SSIA loss in the first few iterations of training.
It also proves that the guidance of macro-perception information 
(in higher layer feature maps) to the lower network layers was the main 
reason for the performance improvement.
Therefore, we use scheme (ii) for the choice of signal sides in all subsequent experiments.


\begin{table}[tb]
  \caption{Comparisons between different attention modules and our SSIA block,
  with ResNet-50 as the baseline.
  Includes the top-1 accuracies (\%) on the CIFAR-10 (C10), CIFAR-100 (C100) and 
  ImageNet-1K (ImageNet) datasets,
  the increments in parameters (+Params), computation (+FLOPs), and memory usage (+MUs)
  compared to the baseline model.
  These baseline model costs shown in the table are for the backbone network 
  (with no classifier head).}
  \label{tab:table3}
  \centering
  \begin{tabular}{l|ccc|c|c|c}
     \toprule
     Model                  & C10            & C100       & ImageNet        & +Params        & +FLOPs        & +MUs \\
     \midrule
     ResNet50 (baseline)    & 93.63          & 74.45           & 76.21           & (23.51M)       & (4.116G)      & (109.68M) \\
     ~+~SE                  & 95.35          & 77.25           & 76.98           & 2.52M          & 0.009G        & 0.12M \\
     ~+~CBAM                & 95.37          & 77.67           & \textbf{77.21}  & 2.53M          & 0.012G        & 0.4M \\
     ~+~BAM                 & 94.86          & 77.28           & 77.07           & 0.36M          & 0.091G        & 3.05M \\
     ~+~ECA                 & 94.55          & 76.16           & 76.75           & 86             & 0.006G        & 0.18M \\
     ~+~SimAM               & 95.65          & 78.42           & 76.91           & \textbf{0}     & 0.001G        & 21.06M \\
     ~\textbf{+~SSIA block} & \textbf{95.69} & \textbf{78.59}  & 77.13           & \textbf{0}     & \textbf{0}    & \textbf{0} \\
     \bottomrule
  \end{tabular}
\end{table}

\subsection{Image Classification}

In this subsection, we compare the performance of the SSIA block to that of 
other attention methods used in CNNs on the CIFAR datasets \cite{Krizhevsky2009LearningML} 
and ImageNet dataset \cite{Russakovsky2015ImageNetLS}, 
including the widely used SE \cite{Hu2018SqueezeandExcitationN}, CBAM \cite{Woo2018CBAMCB},
and BAM \cite{Park2020ASA}; we also compare them to the lightweight attention methods 
efficient channel attention (ECA) \cite{Wang2020ECANetEC} and SimAM \cite{Yang2021SimAMAS}.
Note that on ImageNet-1K dataset, the performance of our SSIA block was related to 
the batch size (see the Appendix for details), we report the results 
with a batch size of 64. The attention methods we compared were trained under 
the same settings.
As shown in \tab{\ref{tab:table3}}, our implicit attention module,
the SSIA block, significantly improved the performance of the baseline model,
it takes $\sim$$4.2\%$ accuracy improvement on the CIFAR-100 dataset,
even surpassing all other explicit attention methods. On ImageNet-1K dataset,
SSIA block also effectively improves model performance (by nearly $1\%$ accuracy improvement) 
and outperforms most of the state-of-the-art attention methods we compared.

Besides achieving a considerable performance improvement, the SSIA block has 
the following two additional advantages:
(1) it generates more substantial and direct supervisory signals to guide 
attention by building self-supervised learning tasks without designing complex 
procedures or requiring additional manual annotation, and 
(2) it works entirely during the training phase and does not introduce any 
parameters, computation, or memory access costs at the inference phase.

Since many attention modules are designed to be very lightweight 
with few additional parameters and computational costs (such as ECA and SimAM),
the second advantage of SSIA block may not seem obvious.
However, in practice, an explicit attention method always introduces branching structures,
they introduce a much larger inference time cost than intuitively we think.
How branching structures affect the inference speed of a model has been discussed 
in detail in recent works (\eg RepVGG \cite{Ding2021RepVGGMV}) that attempt to eliminate 
branching structures in networks.

\subsection{Visualization Results}

\begin{figure*}[tb]
  \centering
  \includegraphics[width=0.99\linewidth]{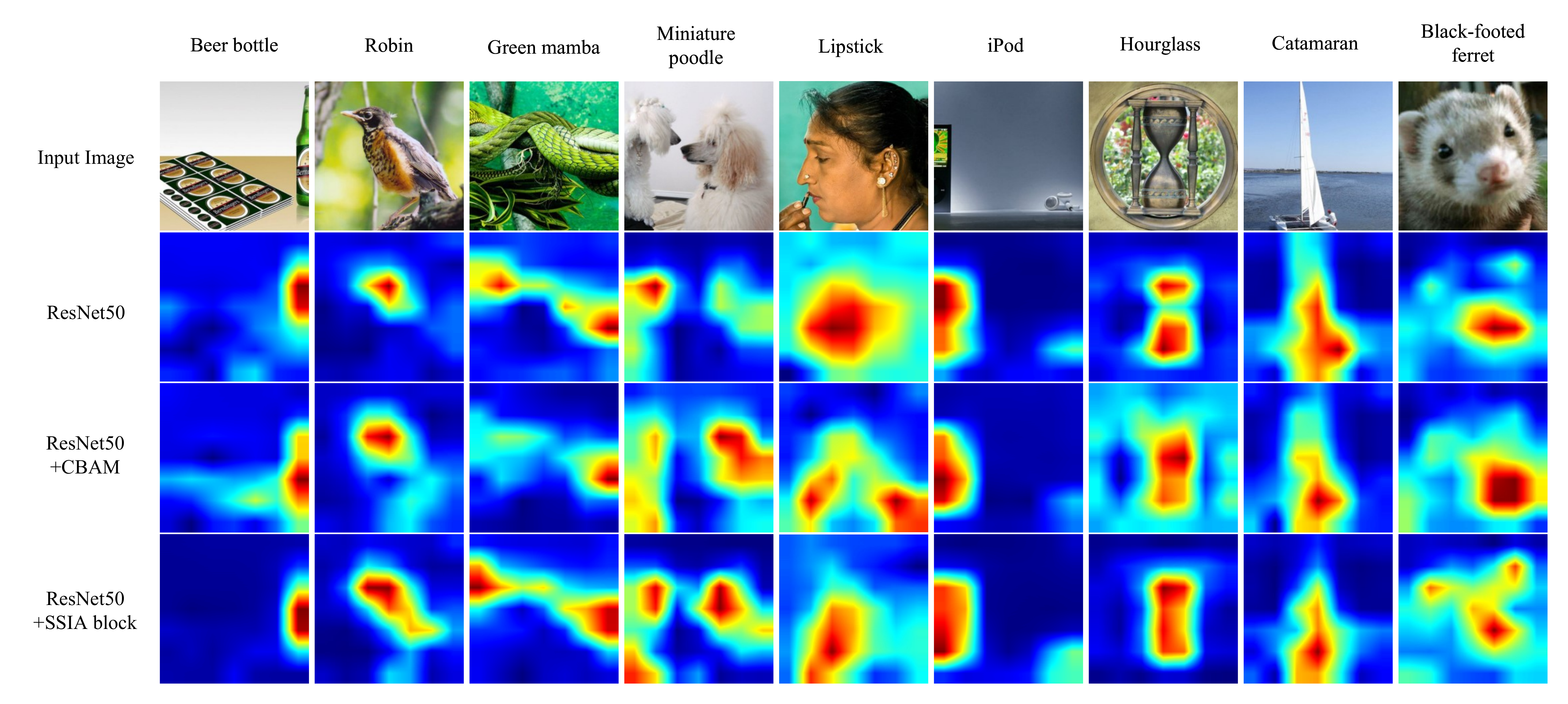}\\
  \caption{CAM visualization results. This visualization compares 
  the CAM \cite{Zhou2016LearningDF} for ResNet-50 
  trained with vanilla, CBAM \cite{Woo2018CBAMCB}, and our SSIA block.
  The CAM visualization is calculated for the last convolutional layer outputs.}
  \label{fig:figure5}
\end{figure*}

To verify whether the SSIA block helped the baseline model acquire better 
attentional capabilities, we used class activation mapping (CAM) \cite{Zhou2016LearningDF} 
to visualize the results.
As shown in \fig{\ref{fig:figure5}}, the CAM activation of the SSIA-block-integrated 
model produced more significant attentional properties than the vanilla model.

It can be seen in most cases that the model integrated with the SSIA block
more effectively suppressed the background and covered a more complete area of the 
main object regions, which indicates that the model has better attentional properties.
Because the SSIA block does not work during inference, we argue that 
these benefits are due to the SSIA block guiding the baseline model to obtain explicit 
attention (\ie the SSIA attention) during training.
Moreover, this suggests that specialized attention modules are \textbf{not} 
a necessary component for implementing attention mechanisms.

\section{Conclusion}

In this paper, we review the implementation of existing attention mechanisms 
and propose SSIA, a novel implementation paradigm for attention mechanism.
We further explore a particular implementation, called the SSIA block,
for implementing SSIA in CNN models.
The SSIA block plugged into the baseline model can simply be discarded after the
training phase, so it does not introduce any costs during the inference phase.
Our experimental results showed that the SSIA block works well in visual 
classification tasks and is very competitive with other well-known
explicit attention methods used in CNNs.
Our visualization results proved that the baseline model integrated with an 
SSIA block has better attentional properties.
In future work, we plan to investigate various SSIA implementations 
and use them in other computer-vision tasks.
To the best of our knowledge, this paper is the first to propose the use of 
implicit attention instead of explicit attention.
We hope that our work will inspire new studies on the attention methods 
used in deep-learning models.


{
\small
\bibliographystyle{plainnat}
\bibliography{egbib}

\begin{thebibliography}{49}
\providecommand{\natexlab}[1]{#1}
\providecommand{\url}[1]{\texttt{#1}}
\expandafter\ifx\csname urlstyle\endcsname\relax
  \providecommand{\doi}[1]{doi: #1}\else
  \providecommand{\doi}{doi: \begingroup \urlstyle{rm}\Url}\fi

\bibitem[Cao et~al.(2019)Cao, Xu, Lin, Wei, and Hu]{Cao2019GCNetNN}
Yue Cao, Jiarui Xu, Stephen Lin, Fangyun Wei, and Han Hu.
\newblock Gcnet: Non-local networks meet squeeze-excitation networks and
  beyond.
\newblock \emph{2019 IEEE/CVF International Conference on Computer Vision
  Workshop (ICCVW)}, pages 1971--1980, 2019.

\bibitem[Chen et~al.(2018{\natexlab{a}})Chen, Papandreou, Kokkinos, Murphy, and
  Yuille]{Chen2018DeepLabSI}
Liang-Chieh Chen, George Papandreou, Iasonas Kokkinos, Kevin~P. Murphy, and
  Alan~Loddon Yuille.
\newblock Deeplab: Semantic image segmentation with deep convolutional nets,
  atrous convolution, and fully connected crfs.
\newblock \emph{IEEE Transactions on Pattern Analysis and Machine
  Intelligence}, 40:\penalty0 834--848, 2018{\natexlab{a}}.

\bibitem[Chen and He(2021)]{Chen2021ExploringSS}
Xinlei Chen and Kaiming He.
\newblock Exploring simple siamese representation learning.
\newblock \emph{2021 IEEE/CVF Conference on Computer Vision and Pattern
  Recognition (CVPR)}, pages 15745--15753, 2021.

\bibitem[Chen et~al.(2018{\natexlab{b}})Chen, Kalantidis, Li, Yan, and
  Feng]{Chen2018A2NetsDA}
Yunpeng Chen, Yannis Kalantidis, Jianshu Li, Shuicheng Yan, and Jiashi Feng.
\newblock A2-nets: Double attention networks.
\newblock In \emph{NeurIPS}, 2018{\natexlab{b}}.

\bibitem[Chollet(2017)]{Chollet2017XceptionDL}
François Chollet.
\newblock Xception: Deep learning with depthwise separable convolutions.
\newblock \emph{2017 IEEE Conference on Computer Vision and Pattern Recognition
  (CVPR)}, pages 1800--1807, 2017.

\bibitem[Ding et~al.(2021)Ding, Zhang, Ma, Han, Ding, and
  Sun]{Ding2021RepVGGMV}
Xiaohan Ding, X.~Zhang, Ningning Ma, Jungong Han, Guiguang Ding, and Jian Sun.
\newblock Repvgg: Making vgg-style convnets great again.
\newblock \emph{2021 IEEE/CVF Conference on Computer Vision and Pattern
  Recognition (CVPR)}, pages 13728--13737, 2021.

\bibitem[Donahue et~al.(2017)Donahue, Kr{\"a}henb{\"u}hl, and
  Darrell]{Donahue2017AdversarialFL}
Jeff Donahue, Philipp Kr{\"a}henb{\"u}hl, and Trevor Darrell.
\newblock Adversarial feature learning.
\newblock \emph{ArXiv}, abs/1605.09782, 2017.

\bibitem[Fukui et~al.(2019)Fukui, Hirakawa, Yamashita, and
  Fujiyoshi]{Fukui2019AttentionBN}
Hiroshi Fukui, Tsubasa Hirakawa, Takayoshi Yamashita, and Hironobu Fujiyoshi.
\newblock Attention branch network: Learning of attention mechanism for visual
  explanation.
\newblock \emph{2019 IEEE/CVF Conference on Computer Vision and Pattern
  Recognition (CVPR)}, pages 10697--10706, 2019.

\bibitem[Gidaris et~al.(2018)Gidaris, Singh, and
  Komodakis]{Gidaris2018UnsupervisedRL}
Spyros Gidaris, Praveer Singh, and Nikos Komodakis.
\newblock Unsupervised representation learning by predicting image rotations.
\newblock \emph{ArXiv}, abs/1803.07728, 2018.

\bibitem[Glorot and Bengio(2010)]{Glorot2010UnderstandingTD}
Xavier Glorot and Yoshua Bengio.
\newblock Understanding the difficulty of training deep feedforward neural
  networks.
\newblock In \emph{AISTATS}, 2010.

\bibitem[Goodfellow et~al.(2015)Goodfellow, Bengio, and
  Courville]{Goodfellow2015DeepL}
Ian~J. Goodfellow, Yoshua Bengio, and Aaron~C. Courville.
\newblock Deep learning.
\newblock \emph{Nature}, 521:\penalty0 436--444, 2015.

\bibitem[Grill et~al.(2020)Grill, Strub, Altch'e, Tallec, Richemond,
  Buchatskaya, Doersch, Pires, Guo, Azar, Piot, Kavukcuoglu, Munos, and
  Valko]{Grill2020BootstrapYO}
Jean-Bastien Grill, Florian Strub, Florent Altch'e, Corentin Tallec, Pierre~H.
  Richemond, Elena Buchatskaya, Carl Doersch, Bernardo~{\'A}vila Pires,
  Zhaohan~Daniel Guo, Mohammad~Gheshlaghi Azar, Bilal Piot, Koray Kavukcuoglu,
  R{\'e}mi Munos, and Michal Valko.
\newblock Bootstrap your own latent: A new approach to self-supervised
  learning.
\newblock \emph{ArXiv}, abs/2006.07733, 2020.

\bibitem[He et~al.(2015)He, Zhang, Ren, and Sun]{He2015DelvingDI}
Kaiming He, X.~Zhang, Shaoqing Ren, and Jian Sun.
\newblock Delving deep into rectifiers: Surpassing human-level performance on
  imagenet classification.
\newblock \emph{2015 IEEE International Conference on Computer Vision (ICCV)},
  pages 1026--1034, 2015.

\bibitem[He et~al.(2016)He, Zhang, Ren, and Sun]{He2016DeepRL}
Kaiming He, X.~Zhang, Shaoqing Ren, and Jian Sun.
\newblock Deep residual learning for image recognition.
\newblock \emph{2016 IEEE Conference on Computer Vision and Pattern Recognition
  (CVPR)}, pages 770--778, 2016.

\bibitem[He et~al.(2019)He, Zhang, Zhang, Zhang, Xie, and Li]{He2019BagOT}
Tong He, Zhi Zhang, Hang Zhang, Zhongyue Zhang, Junyuan Xie, and Mu~Li.
\newblock Bag of tricks for image classification with convolutional neural
  networks.
\newblock \emph{2019 IEEE/CVF Conference on Computer Vision and Pattern
  Recognition (CVPR)}, pages 558--567, 2019.

\bibitem[Hu et~al.(2018{\natexlab{a}})Hu, Shen, Albanie, Sun, and
  Vedaldi]{Hu2018GatherExciteEF}
Jie Hu, Li~Shen, Samuel Albanie, Gang Sun, and Andrea Vedaldi.
\newblock Gather-excite: Exploiting feature context in convolutional neural
  networks.
\newblock In \emph{NeurIPS}, 2018{\natexlab{a}}.

\bibitem[Hu et~al.(2018{\natexlab{b}})Hu, Shen, and
  Sun]{Hu2018SqueezeandExcitationN}
Jie Hu, Li~Shen, and Gang Sun.
\newblock Squeeze-and-excitation networks.
\newblock \emph{2018 IEEE/CVF Conference on Computer Vision and Pattern
  Recognition}, pages 7132--7141, 2018{\natexlab{b}}.

\bibitem[Huang et~al.(2017)Huang, Liu, and Weinberger]{Huang2017DenselyCC}
Gao Huang, Zhuang Liu, and Kilian~Q. Weinberger.
\newblock Densely connected convolutional networks.
\newblock \emph{2017 IEEE Conference on Computer Vision and Pattern Recognition
  (CVPR)}, pages 2261--2269, 2017.

\bibitem[Kingma and Ba(2015)]{Kingma2015AdamAM}
Diederik~P. Kingma and Jimmy Ba.
\newblock Adam: A method for stochastic optimization.
\newblock \emph{CoRR}, abs/1412.6980, 2015.

\bibitem[Krizhevsky(2009)]{Krizhevsky2009LearningML}
Alex Krizhevsky.
\newblock Learning multiple layers of features from tiny images.
\newblock In \emph{Technical report, University of Toronto}, 2009.

\bibitem[Krogh and Hertz(1991)]{Krogh1991ASW}
Anders Krogh and John~A. Hertz.
\newblock A simple weight decay can improve generalization.
\newblock In \emph{NIPS}, 1991.

\bibitem[Li and Chen(2020)]{Li2020DeepRA}
Duo Li and Qifeng Chen.
\newblock Deep reinforced attention learning for quality-aware visual
  recognition.
\newblock In \emph{ECCV}, 2020.

\bibitem[Lin et~al.(2014)Lin, Maire, Belongie, Hays, Perona, Ramanan,
  Doll{\'a}r, and Zitnick]{Lin2014MicrosoftCC}
Tsung-Yi Lin, Michael Maire, Serge~J. Belongie, James Hays, Pietro Perona, Deva
  Ramanan, Piotr Doll{\'a}r, and C.~Lawrence Zitnick.
\newblock Microsoft coco: Common objects in context.
\newblock In \emph{ECCV}, 2014.

\bibitem[Linsley et~al.(2019)Linsley, Shiebler, Eberhardt, and
  Serre]{Linsley2019LearningWA}
Drew Linsley, Dan Shiebler, Sven Eberhardt, and Thomas Serre.
\newblock Learning what and where to attend.
\newblock In \emph{ICLR}, 2019.

\bibitem[Noroozi and Favaro(2016)]{Noroozi2016UnsupervisedLO}
Mehdi Noroozi and Paolo Favaro.
\newblock Unsupervised learning of visual representations by solving jigsaw
  puzzles.
\newblock In \emph{ECCV}, 2016.

\bibitem[Noroozi et~al.(2017)Noroozi, Pirsiavash, and
  Favaro]{Noroozi2017RepresentationLB}
Mehdi Noroozi, Hamed Pirsiavash, and Paolo Favaro.
\newblock Representation learning by learning to count.
\newblock \emph{2017 IEEE International Conference on Computer Vision (ICCV)},
  pages 5899--5907, 2017.

\bibitem[Park et~al.(2020)Park, Woo, Lee, and Kweon]{Park2020ASA}
Jongchan Park, Sanghyun Woo, Joon-Young Lee, and In~So Kweon.
\newblock A simple and light-weight attention module for convolutional neural
  networks.
\newblock \emph{International Journal of Computer Vision}, 128:\penalty0
  783--798, 2020.

\bibitem[Paszke et~al.(2019)Paszke, Gross, Massa, Lerer, Bradbury, Chanan,
  Killeen, Lin, Gimelshein, Antiga, Desmaison, K{\"o}pf, Yang, DeVito, Raison,
  Tejani, Chilamkurthy, Steiner, Fang, Bai, and Chintala]{Paszke2019PyTorchAI}
Adam Paszke, Sam Gross, Francisco Massa, Adam Lerer, James Bradbury, Gregory
  Chanan, Trevor Killeen, Zeming Lin, Natalia Gimelshein, Luca Antiga, Alban
  Desmaison, Andreas K{\"o}pf, Edward Yang, Zach DeVito, Martin Raison, Alykhan
  Tejani, Sasank Chilamkurthy, Benoit Steiner, Lu~Fang, Junjie Bai, and Soumith
  Chintala.
\newblock Pytorch: An imperative style, high-performance deep learning library.
\newblock In \emph{NeurIPS}, 2019.

\bibitem[Patro and Namboodiri(2021)]{Patro2021SelfSF}
Badri~N. Patro and Vinay~P. Namboodiri.
\newblock Self supervision for attention networks.
\newblock \emph{2021 IEEE Winter Conference on Applications of Computer Vision
  (WACV)}, pages 726--735, 2021.

\bibitem[Russakovsky et~al.(2015)Russakovsky, Deng, Su, Krause, Satheesh, Ma,
  Huang, Karpathy, Khosla, Bernstein, Berg, and
  Fei-Fei]{Russakovsky2015ImageNetLS}
Olga Russakovsky, Jia Deng, Hao Su, Jonathan Krause, Sanjeev Satheesh, Sean Ma,
  Zhiheng Huang, Andrej Karpathy, Aditya Khosla, Michael~S. Bernstein,
  Alexander~C. Berg, and Li~Fei-Fei.
\newblock Imagenet large scale visual recognition challenge.
\newblock \emph{International Journal of Computer Vision}, 115:\penalty0
  211--252, 2015.

\bibitem[Sandler et~al.(2018)Sandler, Howard, Zhu, Zhmoginov, and
  Chen]{Sandler2018MobileNetV2IR}
Mark Sandler, Andrew~G. Howard, Menglong Zhu, Andrey Zhmoginov, and Liang-Chieh
  Chen.
\newblock Mobilenetv2: Inverted residuals and linear bottlenecks.
\newblock \emph{2018 IEEE/CVF Conference on Computer Vision and Pattern
  Recognition}, pages 4510--4520, 2018.

\bibitem[Simonyan and Zisserman(2015)]{Simonyan2015VeryDC}
Karen Simonyan and Andrew Zisserman.
\newblock Very deep convolutional networks for large-scale image recognition.
\newblock \emph{CoRR}, abs/1409.1556, 2015.

\bibitem[Srivastava et~al.(2014)Srivastava, Hinton, Krizhevsky, Sutskever, and
  Salakhutdinov]{Srivastava2014DropoutAS}
Nitish Srivastava, Geoffrey~E. Hinton, Alex Krizhevsky, Ilya Sutskever, and
  Ruslan Salakhutdinov.
\newblock Dropout: a simple way to prevent neural networks from overfitting.
\newblock \emph{J. Mach. Learn. Res.}, 15:\penalty0 1929--1958, 2014.

\bibitem[Szegedy et~al.(2016)Szegedy, Vanhoucke, Ioffe, Shlens, and
  Wojna]{Szegedy2016RethinkingTI}
Christian Szegedy, Vincent Vanhoucke, Sergey Ioffe, Jonathon Shlens, and
  Zbigniew Wojna.
\newblock Rethinking the inception architecture for computer vision.
\newblock \emph{2016 IEEE Conference on Computer Vision and Pattern Recognition
  (CVPR)}, pages 2818--2826, 2016.

\bibitem[Vincent et~al.(2010)Vincent, Larochelle, Lajoie, Bengio, and
  Manzagol]{Vincent2010StackedDA}
Pascal Vincent, H.~Larochelle, Isabelle Lajoie, Yoshua Bengio, and
  Pierre-Antoine Manzagol.
\newblock Stacked denoising autoencoders: Learning useful representations in a
  deep network with a local denoising criterion.
\newblock \emph{J. Mach. Learn. Res.}, 11:\penalty0 3371--3408, 2010.

\bibitem[Vinyals et~al.(2016)Vinyals, Blundell, Lillicrap, Kavukcuoglu, and
  Wierstra]{Vinyals2016MatchingNF}
Oriol Vinyals, Charles Blundell, Timothy~P. Lillicrap, Koray Kavukcuoglu, and
  Daan Wierstra.
\newblock Matching networks for one shot learning.
\newblock In \emph{NIPS}, 2016.

\bibitem[Wang et~al.(2017)Wang, Jiang, Qian, Yang, Li, Zhang, Wang, and
  Tang]{Wang2017ResidualAN}
Fei Wang, Mengqing Jiang, Chen Qian, Shuo Yang, Cheng Li, Honggang Zhang,
  Xiaogang Wang, and Xiaoou Tang.
\newblock Residual attention network for image classification.
\newblock \emph{2017 IEEE Conference on Computer Vision and Pattern Recognition
  (CVPR)}, pages 6450--6458, 2017.

\bibitem[Wang et~al.(2020)Wang, Wu, Zhu, Li, Zuo, and Hu]{Wang2020ECANetEC}
Qilong Wang, Banggu Wu, Pengfei Zhu, P.~Li, Wangmeng Zuo, and Qinghua Hu.
\newblock Eca-net: Efficient channel attention for deep convolutional neural
  networks.
\newblock \emph{2020 IEEE/CVF Conference on Computer Vision and Pattern
  Recognition (CVPR)}, pages 11531--11539, 2020.

\bibitem[Wang et~al.(2018)Wang, Girshick, Gupta, and He]{Wang2018NonlocalNN}
X.~Wang, Ross~B. Girshick, Abhinav~Kumar Gupta, and Kaiming He.
\newblock Non-local neural networks.
\newblock \emph{2018 IEEE/CVF Conference on Computer Vision and Pattern
  Recognition}, pages 7794--7803, 2018.

\bibitem[Woo et~al.(2018)Woo, Park, Lee, and Kweon]{Woo2018CBAMCB}
Sanghyun Woo, Jongchan Park, Joon-Young Lee, and In-So Kweon.
\newblock Cbam: Convolutional block attention module.
\newblock In \emph{ECCV}, 2018.

\bibitem[Xie et~al.(2017)Xie, Girshick, Doll{\'a}r, Tu, and
  He]{Xie2017AggregatedRT}
Saining Xie, Ross~B. Girshick, Piotr Doll{\'a}r, Zhuowen Tu, and Kaiming He.
\newblock Aggregated residual transformations for deep neural networks.
\newblock \emph{2017 IEEE Conference on Computer Vision and Pattern Recognition
  (CVPR)}, pages 5987--5995, 2017.

\bibitem[Yang et~al.(2021)Yang, Zhang, Li, and Xie]{Yang2021SimAMAS}
Lingxiao Yang, Ru-Yuan Zhang, Lida Li, and Xiaohua Xie.
\newblock Simam: A simple, parameter-free attention module for convolutional
  neural networks.
\newblock In \emph{ICML}, 2021.

\bibitem[Yosinski et~al.(2015)Yosinski, Clune, Nguyen, Fuchs, and
  Lipson]{Yosinski2015UnderstandingNN}
Jason Yosinski, Jeff Clune, Anh~M Nguyen, Thomas~J. Fuchs, and Hod Lipson.
\newblock Understanding neural networks through deep visualization.
\newblock \emph{ArXiv}, abs/1506.06579, 2015.

\bibitem[Zagoruyko and Komodakis(2016)]{Zagoruyko2016WideRN}
Sergey Zagoruyko and Nikos Komodakis.
\newblock Wide residual networks.
\newblock \emph{ArXiv}, abs/1605.07146, 2016.

\bibitem[Zbontar et~al.(2021)Zbontar, Jing, Misra, LeCun, and
  Deny]{Zbontar2021BarlowTS}
Jure Zbontar, Li~Jing, Ishan Misra, Yann LeCun, and St{\'e}phane Deny.
\newblock Barlow twins: Self-supervised learning via redundancy reduction.
\newblock In \emph{ICML}, 2021.

\bibitem[Zeiler(2012)]{Zeiler2012ADADELTAAA}
Matthew~D. Zeiler.
\newblock Adadelta: An adaptive learning rate method.
\newblock \emph{ArXiv}, abs/1212.5701, 2012.

\bibitem[Zeiler and Fergus(2014)]{Zeiler2014VisualizingAU}
Matthew~D. Zeiler and Rob Fergus.
\newblock Visualizing and understanding convolutional networks.
\newblock In \emph{ECCV}, 2014.

\bibitem[Zhang et~al.(2018)Zhang, Zhou, Lin, and Sun]{Zhang2018ShuffleNetAE}
Xiangyu Zhang, Xinyu Zhou, Mengxiao Lin, and Jian Sun.
\newblock Shufflenet: An extremely efficient convolutional neural network for
  mobile devices.
\newblock \emph{2018 IEEE/CVF Conference on Computer Vision and Pattern
  Recognition}, pages 6848--6856, 2018.

\bibitem[Zhou et~al.(2016)Zhou, Khosla, Lapedriza, Oliva, and
  Torralba]{Zhou2016LearningDF}
Bolei Zhou, Aditya Khosla, {\`A}gata Lapedriza, Aude Oliva, and Antonio
  Torralba.
\newblock Learning deep features for discriminative localization.
\newblock \emph{2016 IEEE Conference on Computer Vision and Pattern Recognition
  (CVPR)}, pages 2921--2929, 2016.

\end{thebibliography}
}

\clearpage
\appendix

\section{Appendix}

\paragraph{MPP Prediction Results.}
\label{prediction_results}

\begin{figure}[htb]
  \centering
  \includegraphics[width=0.9\linewidth]{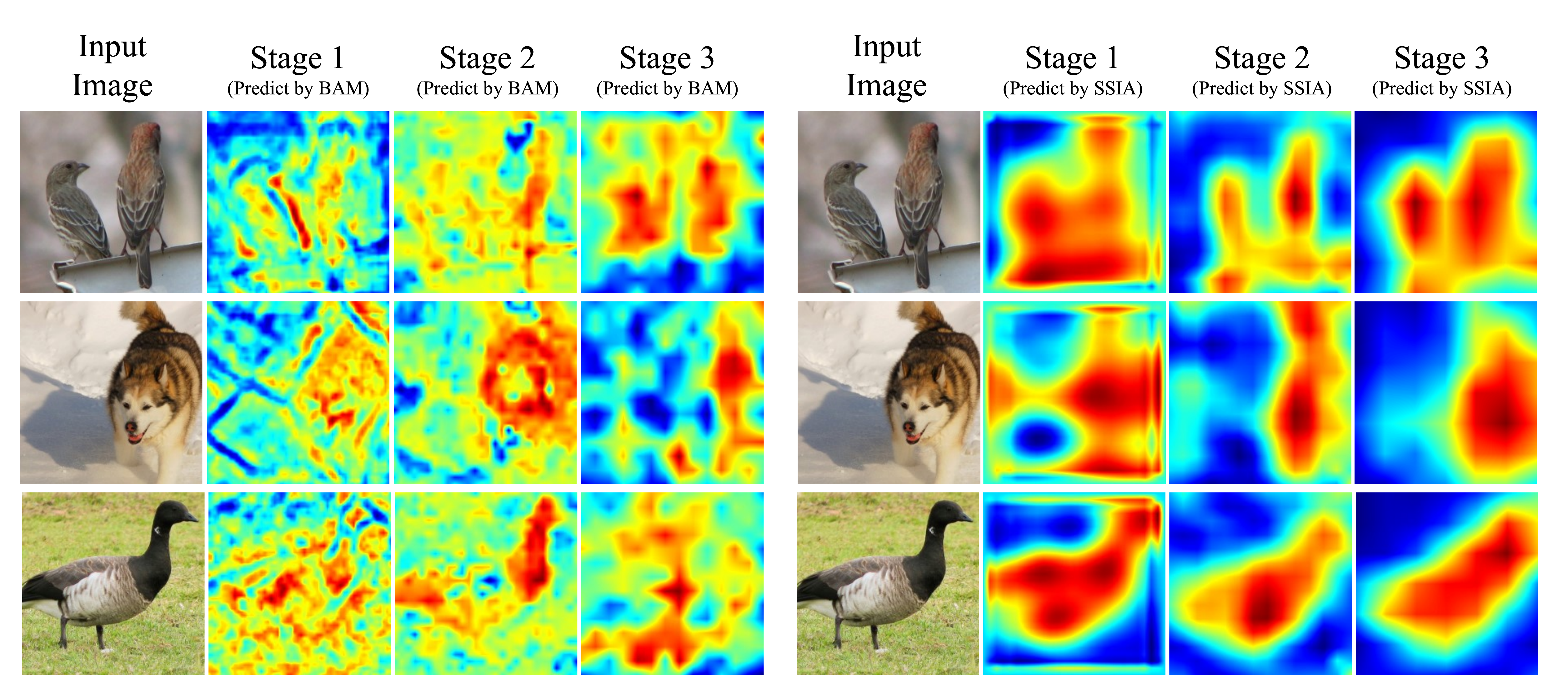}
  \caption{SSIA block and BAM predicted outputs. This visualization 
  used the trained ResNet-50 model with either an integrated SSIA block or BAM \cite{Park2020ASA}.
  The columns from left to right are the input images, the spatial macro-perception 
  predictions (or the spatial attention map predictions) of either the SSIA block or BAM 
  connected at stages 1-3. For the comparison, the results were post-processed and 
  converted to heat maps.}
  \label{fig:figure6}
\end{figure}

We visualized the spatial perception prediction results of the trained MPP,
as shown in \fig{\ref{fig:figure6}}. It can be seen that the macro-perception 
predictions of the SSIA block exhibit strong target and background discrimination capabilities,
with some even made at low network layers.
With existing attention methods such as BAM, whose prediction results are also shown 
in \fig{\ref{fig:figure6}}), it is difficult to distinguish the main object and background 
semantically in the predicted attention maps because these methods are guided only by 
weakly supervised signals. The attention maps predicted by the 
BAM block appear to discriminate the saliency of regions based on texture features,
especially in the lower layers of the network.

Because these results were output as intermediate results in the forward-propagation 
process, a trained MPP could predict global-level semantics directly from the 
current layer feature maps, comparable to that of high-layer feature maps.
These by-products of the SSIA block may be useful for other computer-vision tasks or approaches.

\paragraph{Relationship to Contrastive Learning.}
\label{relation2ssl}

Self-supervised learning is a sub-branch of representation learning,
which aims to mine supervisory information from large-scale unlabeled data 
by pretext tasks \cite{Vincent2010StackedDA, Noroozi2016UnsupervisedLO,
Noroozi2017RepresentationLB, Gidaris2018UnsupervisedRL, Donahue2017AdversarialFL} 
and train the network with that supervised information to learn 
representations that are useful for the target task.
Consider the SSIA block as a self-supervised learning strategy that is not 
designed to learn robust feature representations of images but to make the 
feature representations with better attentional properties (\ie SSIA).
In fact, the core idea of implementing SSIA is to obtain supervisory signals 
by mining the intermediate feature representations of the \emph{previous version} 
of the network rather than relying on annotation information.
These supervisory signals serve as a target to guide the lower layers 
to perform parameter updates, producing the next version of the 
feature representations. That version of the feature representations can 
produce the next guide target in turn, and so forth.
This process is very similar to some contrastive learning methods using 
twin networks, such as \cite{Grill2020BootstrapYO, Chen2021ExploringSS, Zbontar2021BarlowTS}.
To further reveal the subtle relationship between SSIA and the contrastive 
learning paradigm, we have given the Siamese network form of the SSIA block and 
compared it with such contrastive learning methods in the following.

\begin{figure}[tb]
  \centering
  \subfloat[Contrastive learning method]{
     \includegraphics[width=0.50\linewidth]{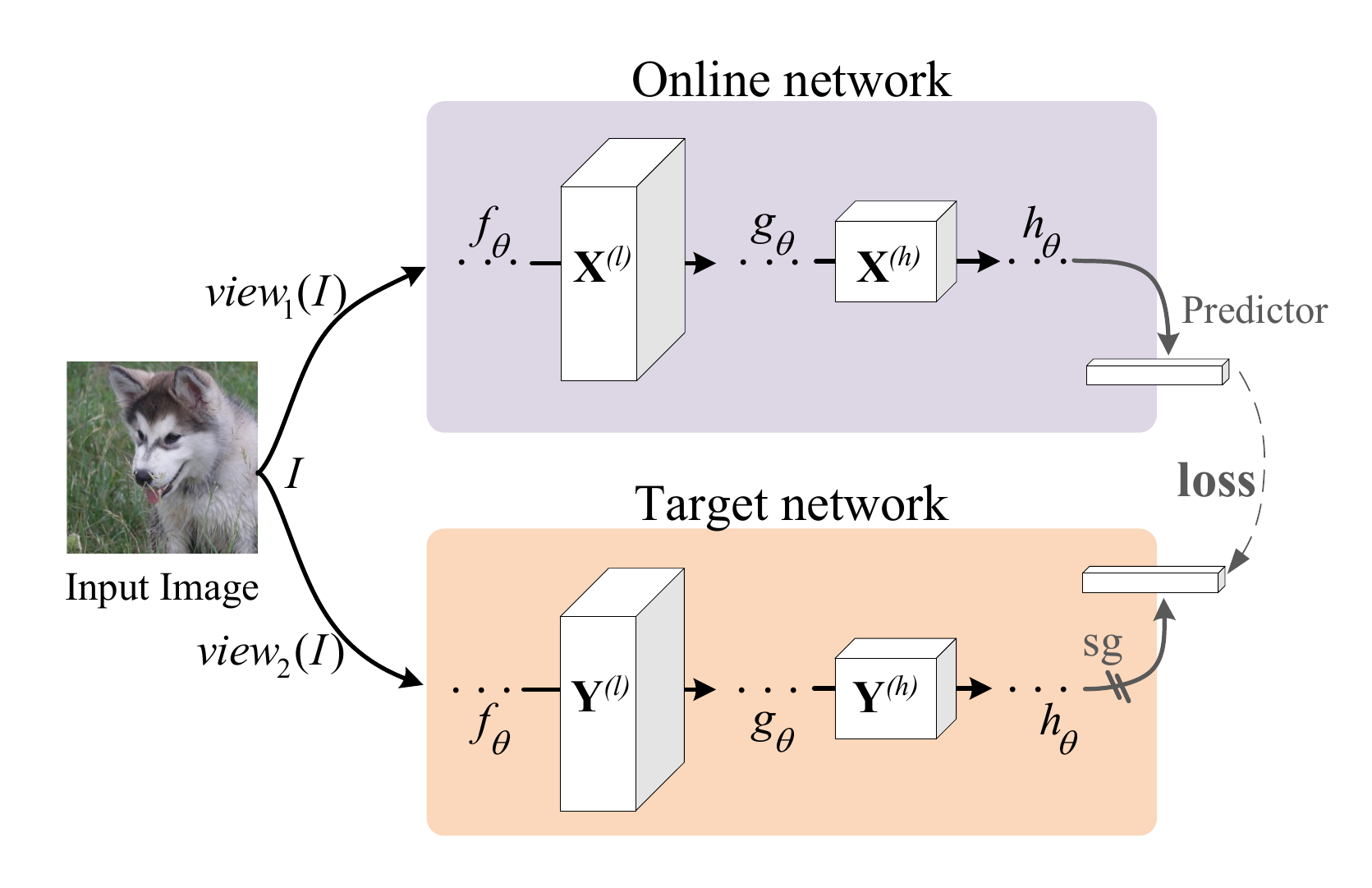}
     \label{fig:figure3:a}
  }
  \subfloat[Siamese network form of the SSIA block]{
     \includegraphics[width=0.46\linewidth]{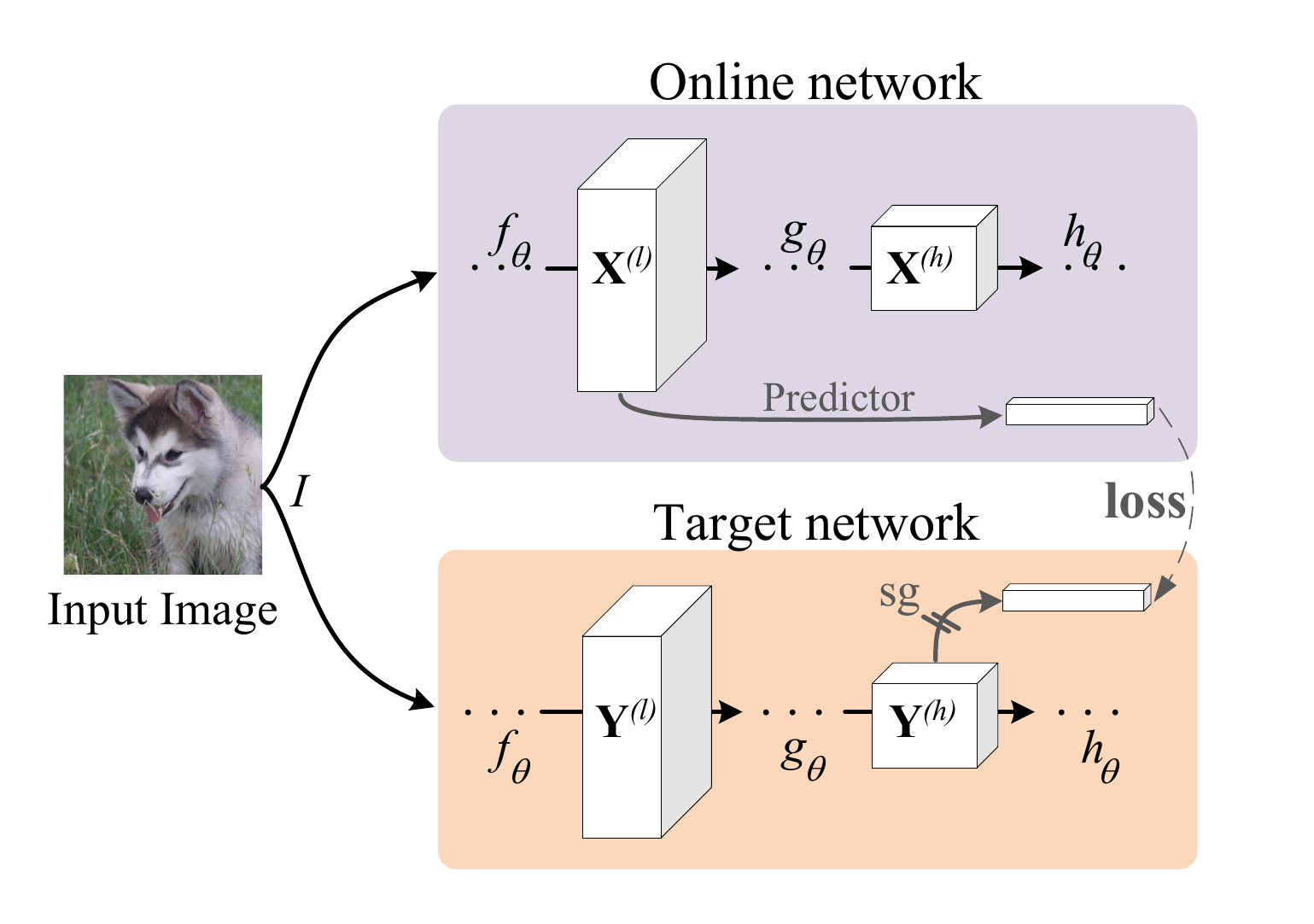}
     \label{fig:figure3:b}
  }\\
  \caption{Comparison on the SSIA block and the contrastive learning paradigm.
  Where (a) is a simplified SimSiam architecture, and (b) is the siamese network 
  form of the SSIA block, "sg" means stop-gradient operation.
  This figure mainly focuses on the process of generating losses from 
  intermediate feature maps, and the network layers are not illustrated 
  for simplicity (denoted by ellipsis dots and letters).}
  \label{fig:figure3}
\end{figure}

\fig{\ref{fig:figure3:a}} shows a simplified version of 
SimSiam \cite{Chen2021ExploringSS}, where the target network is 
a replicated version of the online network (an encoder) sharing 
the same model parameters. Two different augmented views of the 
same input image are formed as a pair of samples as input to 
the Siamese network, and a weak predictor (a MLP) predicts the output vector 
of the target network from the output vector of the online network.
The output vector of the target network is used as supervisory signals 
with a stop-gradient operation, and the weak predictor and the baseline model 
are updated by back-propagation to maximize the similarity between the 
prediction results and the supervisory signals.

\fig{\ref{fig:figure3:b}} shows our proposed SSIA implementation,
the SSIA block, with its Siamese network form implementation, its workflow is 
nearly the same as the contrastive learning workflow in \fig{\ref{fig:figure3:a}},
except 
(1) the online and target networks of the SSIA block have the same input image,
(2) the weak predictor generates macro-perception predictions from the intermediate 
feature map of the online network, and 
(3) the supervisory signals are generated from the intermediate feature map 
of the target network.

It can be seen that the SSIA learning process is similar to the 
representation learning process driven by contrastive learning methods,
as both use the feature representations of the data from the 
previous version of the baseline model during training. Those features 
are used as supervisory signals for updating the model parameters in the next step.
It appears that the SSIA block performs some type of representation learning for 
better attention properties through such a self-supervised paradigm.

Note that the Siamese network form is proposed only for the convenience 
of understanding how SSIA is related to contrastive learning methods,
An SSIA block only forward-propagates once for each sample in practice (only the target network).

\paragraph{Image Classification on ImageNet-1K.}
\label{image_classification_im1k}
On ImageNet-1K dataset, the effectiveness of SSIA block is related to the batch size.
In general, the performance improvement brought from the SSIA block becomes weaker as the 
batch size increases, as shown in \tab{\ref{tab:table5}}. Too large batch size ($\ge$ 256) 
will lead to a weak effect of SSIA block.

\begin{table}[tb]
  \caption{Top-1 accuracy (\%) for ResNet-50 in its vanilla version (baseline) 
  and the SSIA-block-integrated version on ImageNet-1K dataset, with different 
  batch size settings. Both spatial and channel SSIA was used for all SSIA block,
  with the cascaded connection scheme.}
  \label{tab:table5}
  \centering
  \begin{tabular}{l|ccc}
     \toprule
     \multirow{2}{*}{Model}  &      \multicolumn{3}{c}{batch size }          \\
                             & 64             & 128             & 256        \\
     \midrule
     ResNet50 (baseline)    & 76.21          & 76.17           & 76.28       \\
     ~\textbf{+~SSIA block} & \textbf{77.13} & \textbf{76.93}  & 76.41       \\
     \bottomrule
  \end{tabular}
\end{table}

We will investigate the reasons for this phenomenon and its countermeasures 
in future work. Nevertheless, the experiments still confirm that SSIA 
is an effective attention mechanism.

\paragraph{SSIA Block in Different CNN Backbones.}
\label{different_backbone}

We compared the performance of the vanilla version and the SSIA-block-integrated version 
of several popular CNN backbones in \tab{\ref{tab:table4}} on the mini-ImageNet 
\cite{Vinyals2016MatchingNF} dataset,
including VGG-16 \cite{Simonyan2015VeryDC}, ResNet-18/50/101 \cite{He2016DeepRL},
Wide ResNet-18 \cite{Zagoruyko2016WideRN} and ResNeXt-50/101 \cite{Xie2017AggregatedRT}.
All the baseline models with the SSIA block significantly outperformed their vanilla versions,
which suggests the proposed SSIA block can work well on various CNN backbones.

We also applied the SSIA block to the lightweight CNN networks, such as 
MobileNet \cite{Sandler2018MobileNetV2IR} and ShuffleNet \cite{Zhang2018ShuffleNetAE},
as shown in \tab{\ref{tab:table4}}.
Their improvement is not as evident as on other backbone networks, so it appears 
that the SSIA block works especially well on typical CNN networks with residual connections.
However, we found that the SSIA loss declined during training in lightweight networks 
(as in other networks), which implies that the lower network layers indeed learn features 
with better macro-perception information. This may hint us that some design within such 
lightweight networks influences the use of these features in higher network layers,
such as depthwise separable convolutions.

The way to make SSIA attention work well on lightweight networks is a valuable topic 
for implementing deep-learning models in practice, which we intend to investigate 
in the future.

\begin{table}[tb]
  \caption{Top-1 and Top-5 accuracies (\%) for different backbone models in 
  their vanilla version and the integrated-SSIA-block version on the mini-ImageNet
  dataset. We use both spatial and channel SSIA with the cascaded connection scheme
  for all SSIA block.}
  \label{tab:table4}
  \centering
  \begin{tabular}{l|cc|cc}
     \toprule
     \multirow{2}{*}{Backbone} & \multicolumn{2}{c|}{baseline models} & \multicolumn{2}{c}{with SSIA block} \\ 
                               & Top-1 acc. & Top-5 acc.         & Top-1 acc. & Top-5 acc. \\
     \midrule
     VGG16                 & 73.39          & 90.88          & \textbf{74.41} (+1.02)  & \textbf{91.58} (+0.70) \\
     \midrule
     ResNet18              & 78.32          & 92.85          & \textbf{80.19} (+1.87)  & \textbf{93.74} (+0.89) \\
     ResNet50              & 79.59          & 93.45          & \textbf{82.31} (+2.72)  & \textbf{94.65} (+1.20) \\
     ResNet101             & 80.57          & 94.04          & \textbf{82.86} (+2.29)  & \textbf{94.97} (+0.95) \\
     \midrule
     WideResNet18 (widen=1.5)   & 78.78          & 93.16          & \textbf{81.03} (+2.25)  & \textbf{93.95} (+0.79) \\
     WideResNet18 (widen=2.0)   & 79.58          & 93.49          & \textbf{81.60} (+2.02)  & \textbf{94.41} (+0.92) \\
     \midrule
     ResNeXt50  (32x4d)  & 80.25          & 93.79          & \textbf{82.31} (+2.06)  & \textbf{94.76} (+0.97) \\
     ResNeXt101 (32x4d)  & 81.28          & 94.36          & \textbf{82.90} (+1.62)  & \textbf{94.71} (+0.35) \\
     \midrule
     MobileNetV2    & 78.18          & 93.81          & \textbf{78.82} (+0.64)  & \textbf{93.93} (+0.12) \\
     ShuffleNetV2   & 78.33          & 93.59          & \textbf{78.97} (+0.64)  & \textbf{93.75} (+0.16) \\
     \bottomrule
  \end{tabular}
\end{table}

\end{document}